\documentclass[runningheads]{llncs}
\usepackage{graphicx}

\usepackage{tikz}
\usepackage{comment}
\usepackage{amsmath,amssymb} 
\usepackage{color}

\usepackage[accsupp]{axessibility}  

\usepackage{xcolor}
\usepackage{todonotes}
\usepackage[ruled,vlined]{algorithm2e}
\usepackage{multirow}
\usepackage{comment}
\usepackage{hyperref}

\begin{document}
\pagestyle{headings}
\mainmatter
\def\ECCVSubNumber{23}  

\title{Unrestricted Black-box Adversarial Attack Using GAN with Limited Queries} 

\titlerunning{Unrestricted Black-box Adversarial Attack Using GAN with Limited Queries}
%
\author{Dongbin Na \and
Sangwoo Ji \and
Jong Kim}
\authorrunning{Na et al.}
%
\institute{Pohang University of Science and Technology (POSTECH), Pohang, South Korea
\email{\{dongbinna,sangwooji,jkim\}@postech.ac.kr}}
\maketitle

\begin{abstract}
Adversarial examples are inputs intentionally generated for fooling a deep neural network.
Recent studies have proposed unrestricted adversarial attacks that are not norm-constrained.
However, the previous unrestricted attack methods still have limitations to fool real-world applications in a black-box setting.
In this paper, we present a novel method for generating unrestricted adversarial examples using GAN where an attacker can only access the top-1 final decision of a classification model.
Our method, Latent-HSJA, efficiently leverages the advantages of a decision-based attack in the latent space and successfully manipulates the latent vectors for fooling the classification model.

With extensive experiments, we demonstrate that our proposed method is efficient in evaluating the robustness of classification models with limited queries in a black-box setting.
First, we demonstrate that our targeted attack method is query-efficient to produce unrestricted adversarial examples for a facial identity recognition model that contains 307 identities.
Then, we demonstrate that the proposed method can also successfully attack a real-world celebrity recognition service.
The code is available at \textcolor{blue}{\url{https://github.com/ndb796/LatentHSJA}}.
\end{abstract}

\section{Introduction}

\begin{figure}[t]
\begin{center}
   \includegraphics[width=\linewidth]{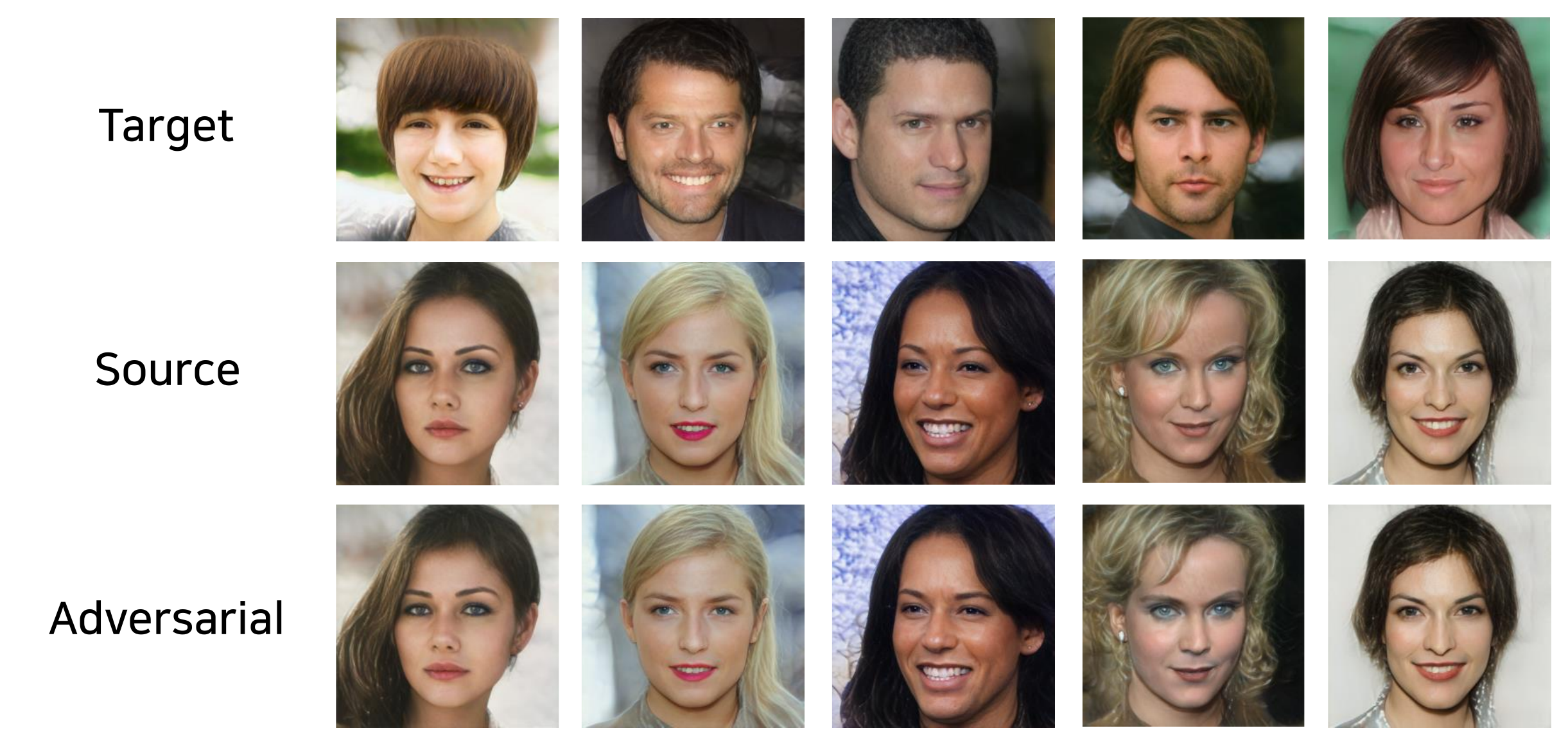}
\end{center}
   \caption{Showcases of our Latent-HSJA attack method against a facial identity recognition model. The first row shows target images and the second row shows source images. The adversarial examples in the last row are classified as target classes and require a feasible number of queries (only 20,000 queries).}
   \label{fig:figure_1}
\end{figure}

Since state-of-the-art deep-learning models have been known to be vulnerable to adversarial attacks~\cite{goodfellow2014explaining,Szegedy}, a large number of defense methods to mitigate the attacks are proposed.
These defense methods include adversarial training~\cite{madry2017towards} and certified defenses~\cite{wong2018provable,cohen2019certified}.
Most previous studies have demonstrated their robustness against adversarial attacks that produce norm-constrained adversarial examples~\cite{madry2017towards,wong2018provable,cohen2019certified}.
The common choices for the constraint are $l_0$, $l_1$, $l_2$, and $l_{\infty}$ norms~\cite{brendel2019accurate}, because a short distance between two image vectors in a image space implies the visual similarity between them.

Recent studies show that adversarial examples can be legitimate even though the perturbation is not small norm-bounded.
These studies have proposed unrestricted adversarial examples that are not norm-constrained but still shown as natural images to humans \cite{poursaeed2019fine,kakizaki2019adversarial,song2018constructing}.
For example, manipulating semantic information such as color schemes or rotation of objects in an image can cause a significant change in the image space while not affecting human perception.
These unrestricted adversarial examples effectively defeat robust models to a norm-constrained perturbation~\cite{song2018constructing,ghiasi2020breaking}.
However, only a few studies~\cite{wang2020amora,kakizaki2019adversarial,song2018constructing} have evaluated the effectiveness of unrestricted adversarial attacks for deep-learning models in a black-box setting.

In a black-box setting, an attacker can only access the output of a classification model.
Real-world applications such as Clarifai and Google Cloud Vision provide only the top-$k$ predictions of the highest confidence scores.
Recent studies have proposed norm-constrained adversarial attack methods for the black-box threat models based on query-response to a classification model \cite{brendel2017decision,ilyas2018black,chen2020hopskipjumpattack}.
However, these black-box attacks have not yet successfully expanded to unrestricted adversarial examples.
Although a few studies have demonstrated their unrestricted adversarial attacks in a black-box setting, their methods suffer from a large number of queries (more than hundreds of thousands of queries) compared with existing norm-based black-box attack methods~\cite{kakizaki2019adversarial} or only support an untargeted attack~\cite{wang2020amora}.

We propose a novel method, Latent-HSJA, for generating unrestricted adversarial examples using GAN in a black-box setting.
To generate unrestricted adversarial examples, we utilize the disentangled style representations of StyleGAN2~\cite{karras2020analyzing}.
Our method manipulates the latent vectors of GAN and efficiently leverages the decision-based attacks in a latent space.
Especially, our targeted attack can be conducted with a target image classified as a target class and a specific source image with limited queries (Figure~\ref{fig:figure_1}).
We mainly deal with the targeted attack because the targeted attack is more difficult to conduct and causes more severe consequences than an untargeted attack.

We show that the proposed method is query-efficient for the targeted attack in a hard-label black-box setting, where an attacker can only access the predicted top-1 label.
It is noted that a black-box attack method should be query-efficient due to the high cost of a query (0.001\$ in Clarifai).
The proposed method for the targeted attack is able to generate an unrestricted adversarial example with a feasible number of queries (less than 20,000 queries) for a facial identity recognition model that contains 307 identities.
This result is comparable to that of state-of-the-art norm-constrained black-box attacks~\cite{ilyas2018black,brendel2017decision,chen2020hopskipjumpattack}.
Especially, our method generates more perceptually superior adversarial examples than previous methods in a super-limited query setting (less than 5,000 queries).
Moreover, we demonstrate that our method can successfully attack a real-world celebrity recognition service.

Our contributions are listed as follows:
\begin{itemize}
\item We propose a query-efficient novel method, Latent-HSJA, for generating unrestricted adversarial examples in a black-box setting. To the best of our knowledge, our method is the first to leverage the targeted unrestricted adversarial attack in a query-limited black-box setting.
\item We demonstrate that the proposed method successfully defeats state-of-the-art deep neural networks such as a gender classification model, a facial identity recognition model, and the real-world celebrity recognition model with a limited query budget.
\end{itemize}
\section{Related Work}

\subsection{Adversarial Examples}

Many deep-learning applications have been deployed in security-important areas such as face recognition \cite{balaban2015deep}, self-driving car \cite{grigorescu2020survey}, and malware detection \cite{Droid}.
However, the recent deep neural network (DNN) models have been known to be vulnerable to adversarial examples \cite{carlini2017towards,goodfellow2014explaining,papernot2016limitations}.
A lot of studies have presented adversarial attack methods in various domains such as image, text, and audio applications \cite{alzantot2018generating,carlini2018audio,ebrahimi2017hotflip}.
The adversarial examples can be also used to improve the robustness of automated authentication systems such as CAPTCHAs \cite{shi2021adversarial,na2020captchas}.

\begin{figure}[h]
  \centering
  \includegraphics[width=\textwidth]{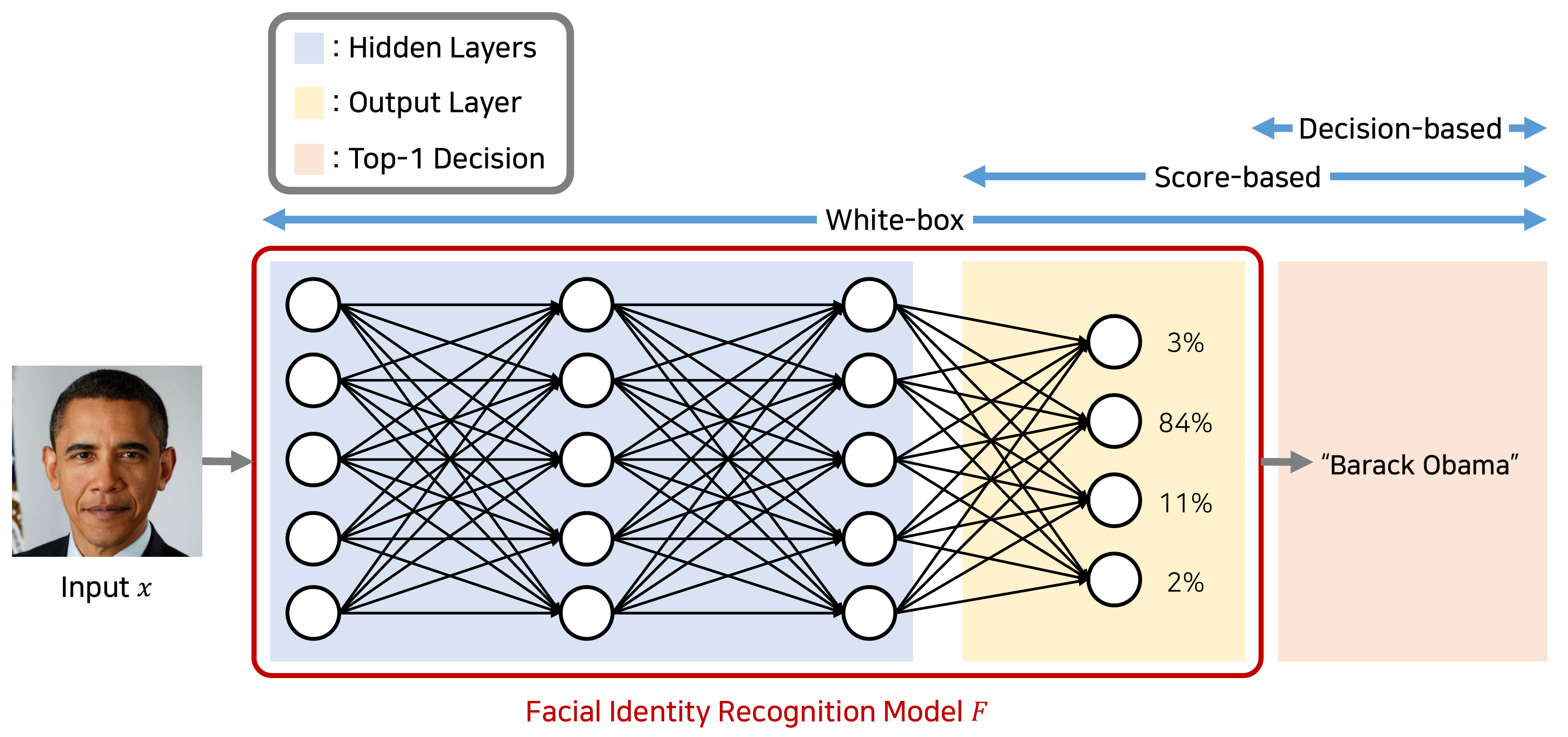}
  \caption{An illustration of the common threat models. In a white-box threat model, the attacker can access the whole information of a DNN model including trained weights. In a score-based threat model, the attacker can access the output layer over all classes. In a decision-based threat model, the attacker can access the top-1 final decision.}
  \label{fig:figure_2}
\end{figure}

One key aspect in categorizing the threat model of adversarial attacks is the accessibility to components of DNN models, known as white-box or black-box (Figure~\ref{fig:figure_2}).
In the white-box threat model, the attacker can access the whole information of a DNN model, including its weights and hyper-parameters.
Many adversarial attacks assume the white-box threat model.
The fast gradient sign method (FGSM) is proposed to generate an adversarial example by calculating the gradient only once \cite{goodfellow2014explaining}.
The projected gradient descent (PGD) is then proposed to efficiently generate a strong adversarial example with several gradient calculation steps \cite{madry2017towards}.
The CW attack is commonly used to find an adversarial example with a small perturbation by calculating a gradient vector typically more than thousands of times \cite{carlini2017towards}.

In the black-box threat model, the attacker can access only the output of a DNN model.
The black-box threat model can be divided into two variants, the score-based threat model and the decision-based threat model.
The score-based threat model assumes that an attacker is able to access the output of the softmax layer of a DNN model.
Natural Evolution Strategy (NES) attack generates adversarial examples by estimating the gradient based on the top-$k$ prediction scores of a DNN model \cite{ilyas2018black}.
On the other hand, the decision-based threat model assumes that an attacker is able to get the final decision of a DNN model, i.e., a predicted label alone.
Boundary Attack (BA) uses random walks along the boundary of a DNN model to generate an adversarial example that looks similar to the source image \cite{brendel2017decision}.
HopSkipJump-Attack (HSJA) is then proposed to produce an adversarial example efficiently by combining binary search and gradient estimation \cite{chen2020hopskipjumpattack}.
As decision-based attacks (BA and HSJA) can access only the top-1 label, they start with an image already classified as the target class and maintain the classification result of the adversarial example during the whole attack procedure \cite{brendel2017decision,chen2020hopskipjumpattack}.

In this paper, we consider the decision-based threat model known as the most difficult black-box setting where an attacker can access only the top-1 label.
This threat model is suitable for a real-world adversarial attack scenario because recent real-world applications such as the Clarifai service may provide only the top-1 label.
Specifically, we present a variant of the HSJA \cite{chen2020hopskipjumpattack}, namely Latent-HSJA, by adding an encoding step to the original HSJA procedure.
The most significant difference between the original HSJA and our attack is that our attack leverages a latent space, not an input image space.

\begin{figure}[h]
  \centering
  \includegraphics[width=240px]{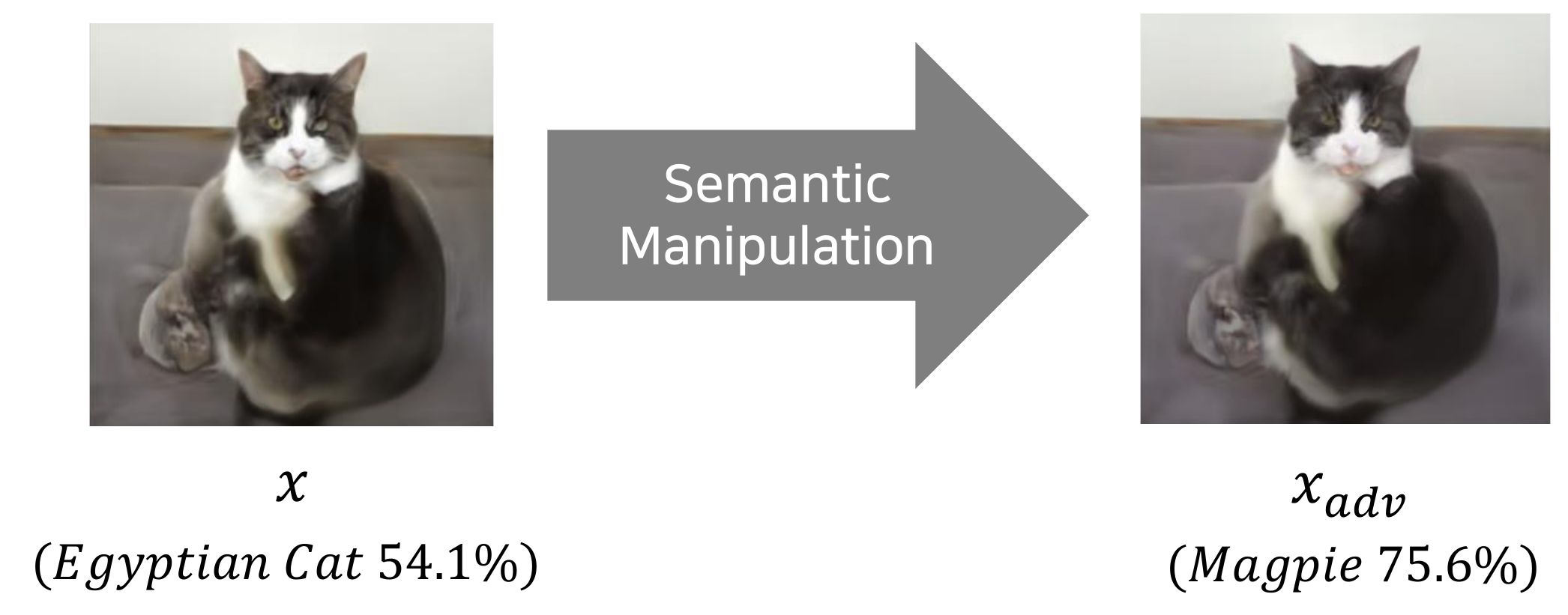}
  \caption{An illustration of an unrestricted adversarial example based on an unrestricted attack method \cite{poursaeed2019fine}.}
  \label{fig:figure_3}
\end{figure}

\subsection{Generative Adversarial Networks}

Generative Adversarial Network (GAN) has been proposed to generate plausible new data examples \cite{goodfellow2014generative}.
Especially, GANs with deep convolutional layers have shown remarkable achievements and are able to produce realistic examples in an image-specific domain \cite{radford2015unsupervised}.
Previous studies with GAN have shown that it is possible to generate high-resolution images up to 1024 $\times$ 1024 resolution in various domains such as the human face, vehicles, and animals \cite{karras2017progressive,karras2019style}.
Recently, StyleGAN architecture shows an outstanding quality of the synthesized image by combining progressive training and the idea of style transfer \cite{karras2019style,karras2020analyzing}.

With the advent of GAN, some previous studies have shown that adversarial examples can exist in the distribution of GAN \cite{poursaeed2019fine,athalye2018obfuscated}.
It implies that an attacker can generate various adversarial examples by manipulating latent vectors of GAN.
From this intuition, we propose a novel method that efficiently utilizes GAN for generating unrestricted adversarial examples that look perceptually natural.
With extensive experiments, we have found that the StyleGAN2 architecture is suitable for our Latent-HSJA to efficiently generate realistic unrestricted adversarial examples in a black-box setting \cite{karras2020analyzing}.

\begin{figure*}
  \includegraphics[width=\textwidth]{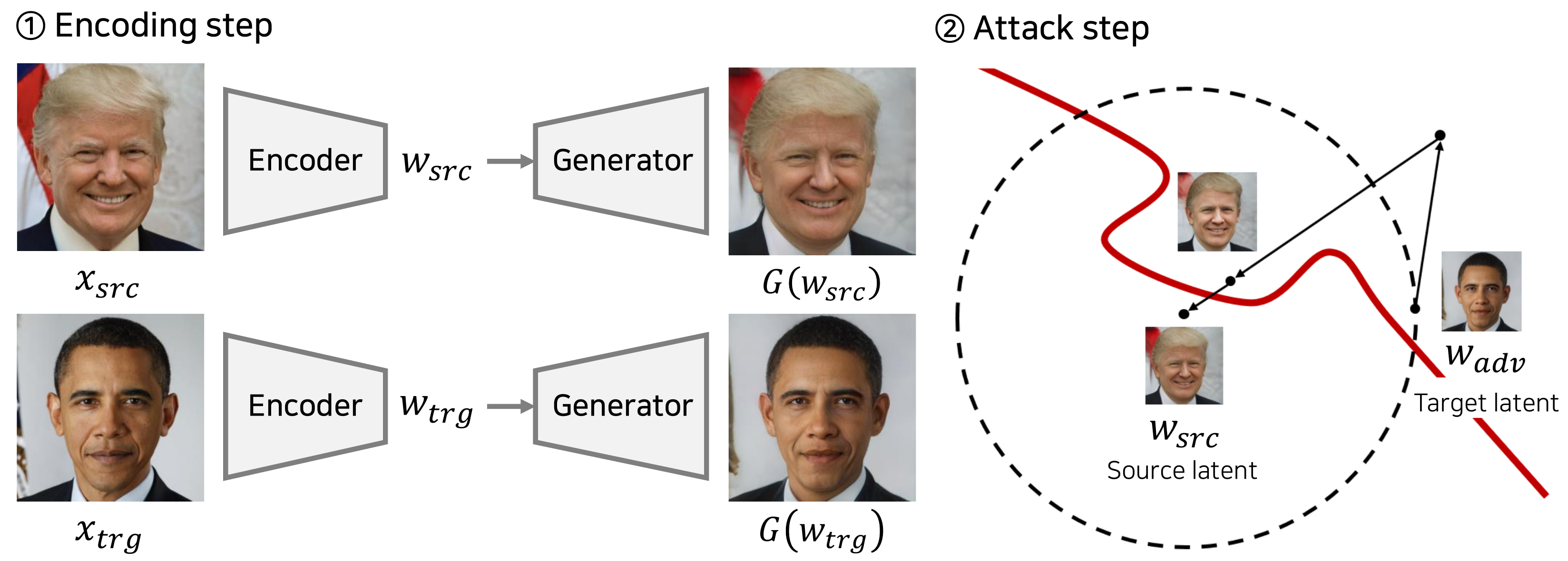}
  \caption{An illustration of our attack method. Our proposed attack method consists of two steps. In the first encoding step, our method predicts two latent vectors ($w_{src}$ and $w_{trg}$) according to input images. In the second attack step, our method drives the latent vector $w_{trg}$ towards the latent vector $w_{src}$ in a latent space. The red line denotes a decision boundary of a classification model in a latent space.}
  \label{fig:figure_4}
\end{figure*}

\subsection{Unrestricted Adversarial Attacks}

Most previous studies consider attack methods that generate adversarial examples constrained to the specific $p$-norm bound \cite{goodfellow2014explaining,carlini2017towards,papernot2016limitations}.
Specifically, $l_0$, $l_1$, $l_2$, and $l_{\infty}$ norms are commonly used ~\cite{brendel2019accurate}.
Therefore, previous defense methods also focus on a norm-constrained adversarial perturbation.
Especially, recently proposed defense methods~\cite{cohen2019certified,wong2018provable} provide certified robustness against an adversarial perturbation with a specific size of $p$-norm bound.
On the other hand, recent studies have proposed various unrestricted adversarial attack methods that are not norm-constrained \cite{poursaeed2019fine,kakizaki2019adversarial,bhattad2019unrestricted,brown2018unrestricted} (Figure~\ref{fig:figure_3}).
Moreover, some studies have shown that these unrestricted attacks can bypass even the certified defense methods \cite{song2018constructing,ghiasi2020breaking}.

Nonetheless, the current unrestricted adversarial attacks have not yet successfully expanded to the black-box threat model.
A related study has proposed an unrestricted black-box attack using an image-to-image translation network.
However, it requires a large number of queries (more than hundreds of thousands of queries) in a black-box setting, which is not desirable for fooling real-world applications \cite{kakizaki2019adversarial}.
In this paper, we present a targeted attack method that generates unrestricted adversarial examples with a limited query budget (less than 20,000 queries).
To the best of our knowledge, we are the first to propose a targeted black-box attack method that generates unrestricted adversarial examples with limited queries.
\section{Proposed Methods}

In a decision-based threat model, the common goal of an attacker is to generate an adversarial example $x_{adv}$ that fools a DNN-based classification model $F(x)$ whose prediction output is the top-1 label for an input $x$ \cite{chen2020hopskipjumpattack,brendel2017decision}.
In the targeted attack setting, the attacker finds an adversarial example $x_{adv}$ that is similar to an $x_{src}$ and is classified as a target class $y_{trg}$ by the model $F$.
The distance metric $D$ is used to minimize the size of an adversarial perturbation.
The common choice of the distance metric $D$ is $p$-norm.

The general form of the objective is as follows:

\begin{equation*}
\begin{aligned}
\underset{x_{adv}}{\text{minimize}}
& & {D(x_{src}, x_{adv})} 
& & \text{s. t. }
& & {F(x_{adv})=y_{trg}}.
\end{aligned}
\end{equation*}

For the unrestricted attack, the attacker sets $D$ as a metric to measure a distance of semantic information such as rotation, hue, saturation, brightness, or high-level styles between two images \cite{bhattad2019unrestricted,ghiasi2020breaking,hosseini2018semantic}.
Our proposed method also minimizes the semantic distance between two images.
We use $D$ as the $l_{2}$ distance between the two latent vectors,  i.e., $w_{src}$ and $w_{adv}$.
We postulate that if the distance between two latent vectors is short enough in the latent space of the GAN model $G$, the two synthesized images are similar in human perception.
Especially when two latent vectors are exactly the same, the images generated by propagating two latent vectors into the GAN model should be the same.

Therefore, our attack uses the following objective:

\begin{equation*}
\begin{aligned}
& & \underset{w_{adv}}{\text{minimize}}
& & {D(w_{src}, w_{adv})} 
& & \text{s. t.}
& & {F(G(w_{adv}))=y_{trg}}. \\
\end{aligned}
\end{equation*}

\subsection{Decision-based Attack in Latent Space}

We propose a method to conduct an unrestricted black-box attack, namely, Latent-HSJA,  consisting of two steps (Figure~\ref{fig:figure_4}).
First, we find latent vectors of source and target images ($w_{src}$ and $w_{trg}$).
We use the latent vector of the target image $w_{trg}$ as an initial latent vector for an adversarial example $w_{adv}$.
The corresponding adversarial example $G(w_{adv})$ should be adversarial (i.e., classified as the target class) at the start of the attack.
Second, we conduct a decision-based update procedure in the latent space.
Our method always maintains the predicted label of the adversarial example to be adversarial during the whole attack procedure ($F(G(w_{adv})) = y_{trg}$).
We illustrate our attack algorithm in Figure~\ref{fig:figure_4}.

We utilize the HSJA method \cite{chen2020hopskipjumpattack} for the second step of the proposed method (the decision-based update).
Our decision-based attack minimizes $D(w_{src}, w_{adv})$ while preserving that the model output $F(G(w_{adv}))$ is always classified as the target class $y_{trg}$.
After we run the attack algorithm, we get an adversarial latent vector $w_{adv}$ such that $G(w_{adv}) \approx x_{src}$ in human perception.
Our method tends to change the semantic information of an adversarial example $G(w_{adv})$ because our method chooses to update the latent vector $w_{adv}$ in the latent space rather than directly update the image $x_{adv}$ in the image space.

\begin{algorithm}[h]
\SetAlgoLined
\textbf{Require:} $Encoder$ denotes an image encoding network, $G$ denotes a pre-trained GAN model, $LatentHSJA$ denotes our attack based on the HopSkipJump-Attack, and $F$ denotes a classification model for the attack.\\
\textbf{Input:} Two input images $x_{src}$, $x_{trg}$.\\
\KwResult{The adversarial example $x_{adv}$.}
  $w_{src} = Encoder(x_{src})$\;
  $w_{trg} = Encoder(x_{trg})$\;
  $w_{adv} = Latent\-HSJA(G, F, w_{src}, w_{trg})$\;
  $x_{adv} = G(w_{adv})$\;
 \caption{Decision-based attack in a latent space}
 \label{algorithm_1}
\end{algorithm}

As a result, we can generate a perceptually natural adversarial example even in an early stage of the attack because our Latent-HSJA updates coarse-grained semantic features of the image.
On the other hand, previous decision-based attack methods based on $p$-norm metrics suffer from a limitation that the generated adversarial example $x_{adv}$ is not perceptually plausible in an early stage of the attack (less than 5,000 queries).

\subsection{Encoding Algorithm}

\begin{table}[t]
\setlength{\tabcolsep}{0.5em}
\begin{center}
\caption{The validation accuracies of our trained models. Our attack method is evaluated on these classification models. \label{table_0}}
\begin{tabular}{|c|c|c|}
\hline
Architectures & Identity Dataset & Gender Dataset \\
\hline
MNasNet1.0 & 78.35\% & 98.38\% \\
\hline
DenseNet121 & 86.42\% & 98.15\% \\
\hline
ResNet18 & 87.82\% & 98.55\% \\
\hline
ResNet101 & 87.98\% & 98.05\% \\
\hline
\end{tabular}
\end{center}
\vspace*{-2mm}
\end{table}

Our attack requires an accurate encoding method that maps an image $x$ into a latent vector $w$ such that $F(x) = F(G(Encoder(x)))$.
Ideally, a perfect encoding method can satisfy this requirement.
We first prepare $w_{trg}$ by using the encoder so that $G(w_{trg})$ is classified as an adversarial class $y_{trg}$.
Secondly, we also prepare $w_{src}$ by using a given source image $x_{src}$.
With this pair ($w_{src}$ and $w_{trg}$), we could finally get the adversarial example $G(w_{adv})$ by conducting the Latent-HSJA (Algorithm~\ref{algorithm_1}).

For developing an encoding method that finds a latent vector according to an image for the StyleGAN-based generators, two approaches are commonly used.
The first is an optimization-based approach that updates a latent vector using gradient descent steps \cite{abdal2019image2stylegan,abdal2020image2stylegan++}.
The second approach trains an additional encoder network that embeds an image to a latent vector \cite{DonahueKD16,richardson2021encoding,tov2021designing}.
Recent studies have also shown that hybrid approaches combining the two methods could return better-encoded results \cite{zhu2020domain,bau2019seeing}.
We have found that the optimization-based encoding method is unsuitable for our attack because it could cause severe overfitting.
Previous studies have also shown that such overfitting is not desirable for semantic image manipulation \cite{zhu2020domain,tov2021designing}.
When the initial latent vector is highly overfitted, our attack could fail.
We have observed that the recent pSp encoder provides successful encoding results for our attack method \cite{richardson2021encoding}.
\section{Experiments}

\subsection{Experiment Settings}

\subsubsection{Dataset}

\begin{figure}[t]
\begin{center}
   \includegraphics[width=1.0\linewidth]{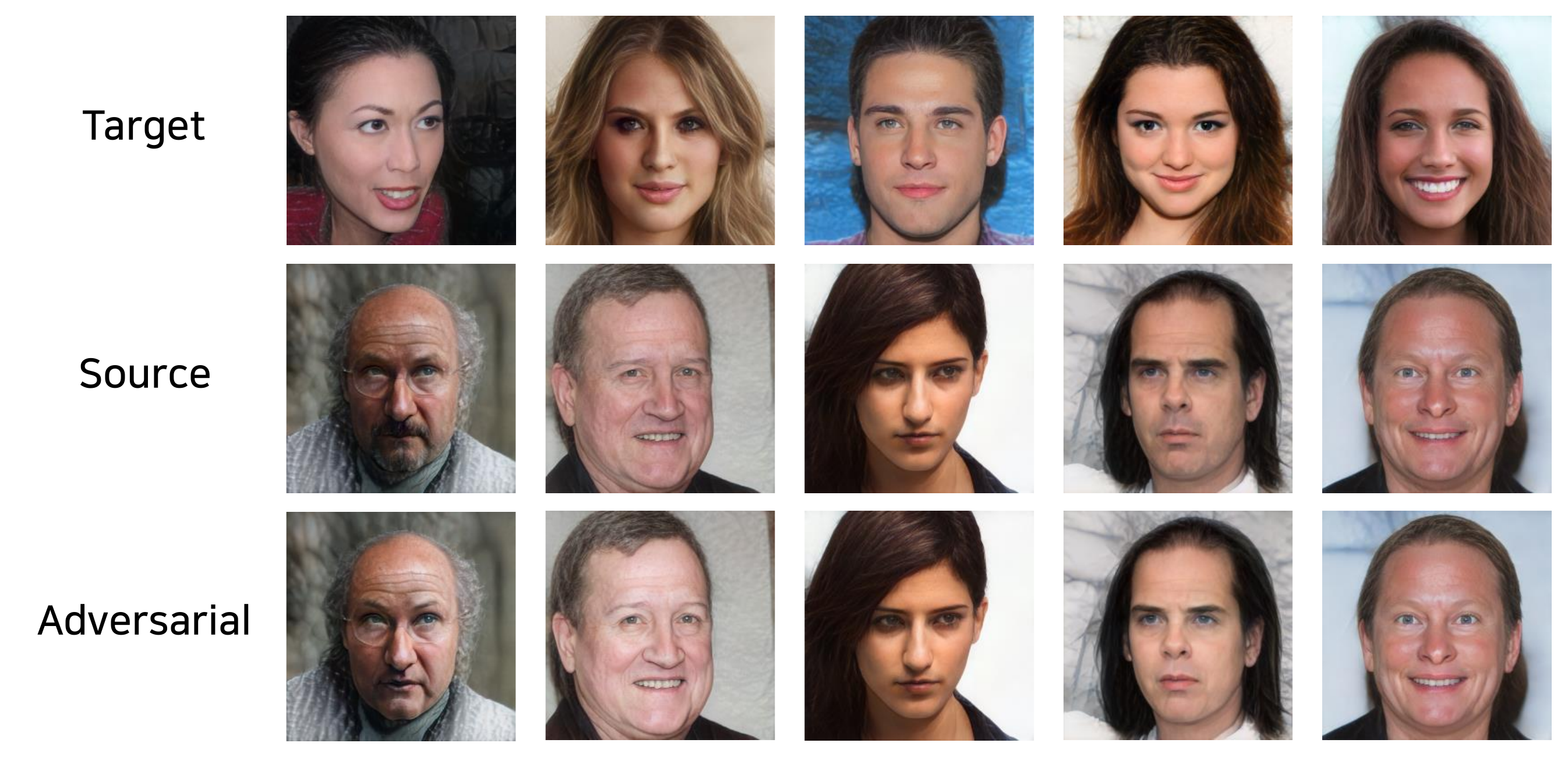}
\end{center}
   \caption{Showcases of our Latent-HSJA attack method against a face gender classification model. The first row shows target images and the second row shows source images. The adversarial examples in the last row are classified as target classes and require a feasible number of queries (only 20,000 queries).}
   \label{fig:figure_5}
\vspace*{-2mm}
\end{figure}

For experiments, we use the CelebA-HQ dataset \cite{CelebAMask-HQ}, a common baseline dataset for face attribute classification tasks.
The CelebA-HQ dataset contains 30,000 face images that are all $1024 \times 1024$ resolution images.
There are 6,217 unique identities and 40 binary attributes in the CelebA-HQ dataset.
First, we filter the CelebA-HQ dataset so that each identity contains more than 15 images for training the facial identity recognition models.
As a result, our filtered facial identity dataset contains 307 identities, and there are 4,263 face images for training and 1,215 face images for validation.
Secondly, we utilize the original CelebA-HQ dataset to train the facial gender classification models.
The CelebA-HQ dataset contains 11,057 male images and 18,943 female images.
We split these two datasets into 4:1 as training and validation.

\subsubsection{Classification Models}

\begin{figure*}
  \includegraphics[width=\textwidth]{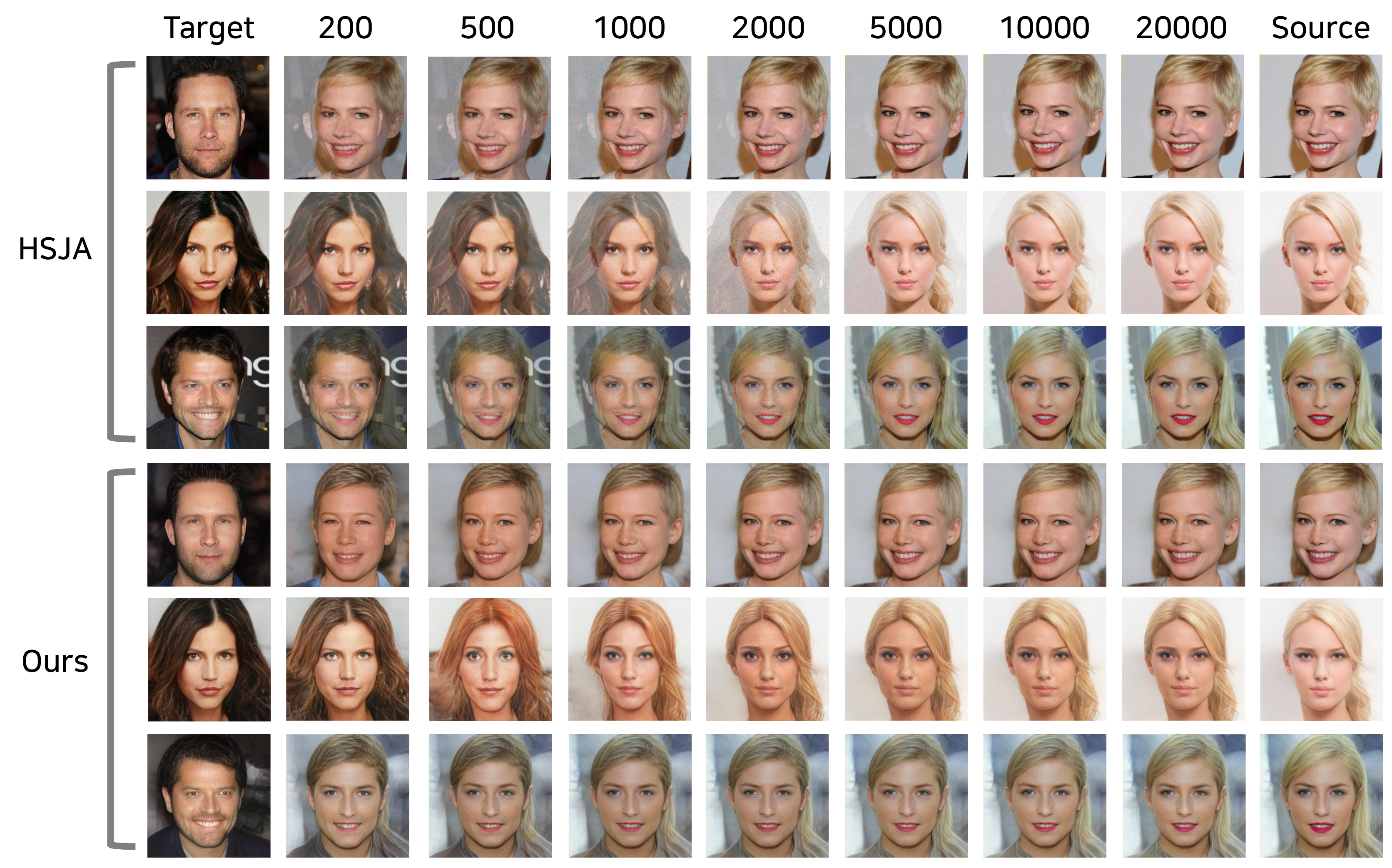}
  \caption{An illustration of our targeted black-box attack results for a facial identity classification model. Our method (Latent-HSJA) is comparable with state-of-the-art norm-constrained adversarial attacks (HSJA) in a black-box setting.}
  \label{fig:figure_6}
\end{figure*}

For validating our Latent-HSJA, we have trained classification models on the two aforementioned datasets.
We have fine-tuned MNasNet1.0, DenseNet121, ResNet18 and ResNet101  \cite{tan2019mnasnet,huang2017densely,he2016deep} that are pre-trained on the ILSVRC2012 dataset.
All classification models resize the resolution of inputs to $256 \times 256$ in the input pre-processing step.
The validation accuracies of all trained models are reported in Table~\ref{table_0}.
We have found the ResNet18 models show good generalization performance for both tasks, thus we report the main experimental results using the fine-tuned ResNet18 models.
We also evaluate our attack method on a real-world application to verify the effectiveness of our method.
The celebrity recognition service of Clarifai contains a large number of facial identities of over 10,000 recognized celebrities.
For each face object of an image, this service returns the top-1 class and its probability.

\subsubsection{Attack Details}

In our attack method, we utilize the StyleGAN2 architecture \cite{karras2020analyzing}.
For updating a encoded latent vector, a previous study utilizes $w^+$ space whose dimension is $18 \times 512$ to get better results \cite{abdal2019image2stylegan}.
Following the previous work, we use $w^+$ space and normalize all the latent vectors so that the values of latent vectors are between [0, 1] and utilize the HSJA in the normalized latent space.
In our experiments, we randomly select 100 (image $x$, encoded latent $w$) pairs such that $F(x)$ is equal to $F(G(w))$ in the validation datasets for facial identity recognition and gender recognition.
As mentioned in the previous section, the goal of our unrestricted attack is to minimize the $l_{2}$ distance between $w_{adv}$ and $w_{src}$.
We note that the attack success rate is always 100\% because Latent-HSJA maintains the adversarial example to be always adversarial in the whole attack procedure, and the objective of attacks is to minimize the similarity distance $D$.
For whole experiments, we report the targeted adversarial attack results.

\subsubsection{Evaluation Metrics}

The adversarial example $x_{adv}$ should be close to the source image $x_{src}$ such that $F(x_{adv}) = y_{trg}$ in the targeted adversarial attack setting.
Therefore, we calculate how different the adversarial example $x_{adv}$ is from the source image $x_{src}$ using several metrics.
We consider the similarity score SIM \cite{huang2020curricularface} and perceptual loss LPIPS \cite{zhang2018unreasonable} for evaluating our attack compared with the previously proposed $p$-norm based attack.
Previous studies have demonstrated that the similarity score and LPIPS can be used for measuring perceptual similarity between two human face images \cite{richardson2021encoding}.

\subsection{Gender Classification}

\begin{table*}[t]
  \setlength{\tabcolsep}{0.5em}
  \begin{center}
    \caption{Experimental results of our Latent-HSJA compared with previous $p$-norm based HSJA against the gender recognition model and identity classification model. All adversarial examples are generated for fooling the ResNet18 models. The SIM and LPIPS scores are calculated using $x_{src}$ and adversarial example $x_{adv}$ ($G(w_{src})$ and $G(w_{adv})$ for Latent-HSJA). \label{table_6}}
    \begin{tabular}{|c|c|c|c|c|c|c|c|}
    \hline
    \multicolumn{2}{|c}{} & \multicolumn{6}{|c|}{Gender Recognition} \\
    \hline
    \multirow{2}{*}{Method} & \multirow{2}{*}{Metric} & \multicolumn{6}{c|}{Model Queries} \\
    \cline{3-8}
    & & 500 & 1000 & 3000 & 5000 & 10000 & 20000 \\
    \hline
    \multirow{2}{*}{HSJA \cite{chen2020hopskipjumpattack}} & SIM↑ & 0.546 & 0.621 & 0.724 & 0.780 & \textbf{0.840} & \textbf{0.886} \\
    \cline{2-8}
    & LPIPS↓ & 0.484 & 0.340 & 0.181 & 0.122 & 0.068 & 0.037 \\
    \hline
    \multirow{2}{*}{Ours} & SIM↑ & \textbf{0.769} & \textbf{0.794} & \textbf{0.817} & \textbf{0.821} & 0.824 & 0.828 \\
    \cline{2-8}
    & LPIPS↓ & \textbf{0.066} & \textbf{0.052} & \textbf{0.041} & \textbf{0.039} & \textbf{0.037} & \textbf{0.036} \\
    \hline
    \multicolumn{2}{|c}{} & \multicolumn{6}{|c|}{Identity Classification} \\
    \hline
    \multirow{2}{*}{Method} & \multirow{2}{*}{Metric} & \multicolumn{6}{c|}{Model Queries} \\
    \cline{3-8}
    & & 500 & 1000 & 3000 & 5000 & 10000 & 20000 \\
    \hline
    \multirow{2}{*}{HSJA \cite{chen2020hopskipjumpattack}} & SIM↑ & 0.538 & 0.569 & 0.663 & 0.728 & \textbf{0.821} & \textbf{0.895} \\
    \cline{2-8}
    & LPIPS↓ & 0.313 & 0.287 & 0.187 & 0.132 & 0.067 & \textbf{0.029} \\
    \hline
    \multirow{2}{*}{Ours} & SIM↑ & \textbf{0.665} & \textbf{0.719} & \textbf{0.779} & \textbf{0.797} & 0.811 & 0.819 \\
    \cline{2-8}
    & LPIPS↓ & \textbf{0.164} & \textbf{0.119} & \textbf{0.072} & \textbf{0.061} & \textbf{0.053} & 0.049 \\
    \hline
    \end{tabular}
  \end{center}
  \vspace*{-2mm}
\end{table*}

\begin{figure}[h]
  \includegraphics[width=\linewidth]{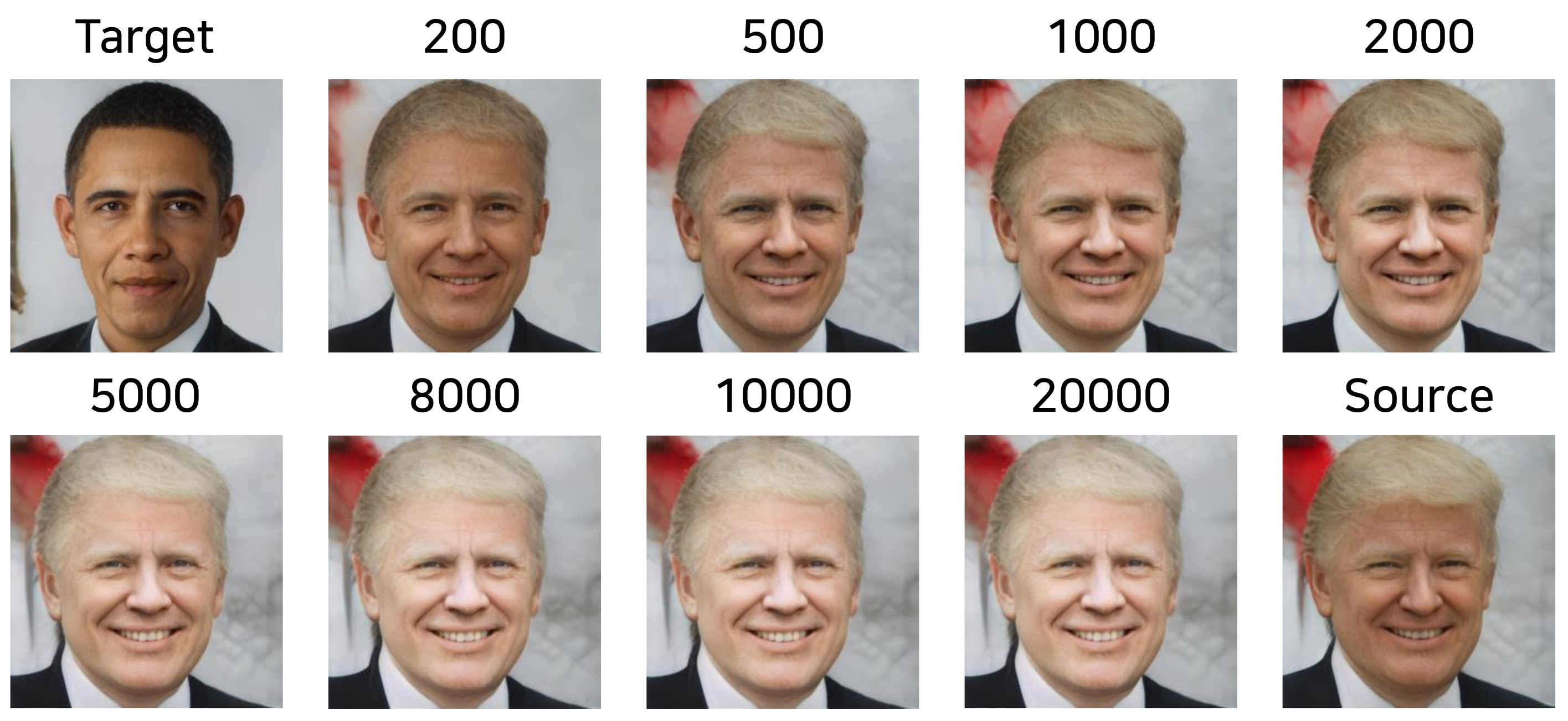}
  \caption{An illustration of our targeted unrestricted black-box attack results for Clarifai service.}
  \label{fig:figure_7}
\end{figure}

\begin{table}[t]
\begin{center}
\setlength{\tabcolsep}{0.5em}
\caption{The results of brute-forcing for finding feasible target image instances that can be used for starting points of our Latent-HSJA using StyleGAN \cite{karras2019style} (higher $\vert Class \vert$ is better). The facial identity classification model contains 307 identities. \label{table_10}}
\begin{tabular}{|c|c||c|c|c|}
\hline
Dataset & Model Queries & $\vert Class \vert$ & $\vert Class \vert_{>50}$ & $\vert Class \vert_{>90}$ \\
\hline
\multirow{5}{*}{FFHQ \cite{karras2019style}} & 1000 & 183 & 117 & 36 \\ \cline{2-5}
 & 5000 & 252 & 203 & 90 \\ \cline{2-5}
 & 10000 & 267 & 241 & 111 \\ \cline{2-5}
 & 40000 & 300 & 275 & 168 \\ \cline{2-5}
 & 80000 & 305 & 288 & 200 \\ \cline{2-5}
\hline
\multirow{5}{*}{CelebA-HQ \cite{CelebAMask-HQ}} & 1000 & 242 & 170 & 46 \\ \cline{2-5}
 & 5000 & 301 & 268 & 131 \\ \cline{2-5}
 & 10000 & 304 & 289 & 171 \\ \cline{2-5}
 & 40000 & 307 & 306 & 245 \\ \cline{2-5}
 & 80000 & 307 & 306 & 274 \\ \cline{2-5}
\hline
\end{tabular}
\end{center}
\vspace*{-2mm}
\end{table}

Gender classification is a common binary classification task for classifying a face image.
For the gender classification model, the targeted attack setting is the same as the untargeted attack setting.
We demonstrate that adversarial examples made from our Latent-HSJA are sufficiently similar to the source images $x_{src}$ (Figure~\ref{fig:figure_5}).
Especially, our method is more efficient in the initial steps than the norm-based adversarial attack (Table~\ref{table_6}).
Our adversarial examples show better results in the evaluation metrics of SIM and LPIPS compared to the norm-based adversarial attack below 5,000 queries.

\subsection{Identity Recognition}

For the evaluation of our method against the facial identity recognition task, we have experimented with the targeted attack.
We demonstrate that our method is query-efficient for the targeted attack.
Our targeted attack is based on Algorithm~\ref{algorithm_1}.
As illustrated in Figure~\ref{fig:figure_6}, our method mainly changes the coarse-grained semantic features in the initial steps (Table~\ref{table_6}).
This property is desirable for the black-box attack since our adversarial example will quickly be semantically far away from the target image.
Moreover, the previous decision-based attacks often generate disrupted images that contain perceptible artificial noises with a limited query budget (under 5,000 queries).
In contrast, our method maintains the adversarial examples to be always perceptually feasible.

\subsection{Real-world Application}

\subsubsection{Ethical Considerations}

We are aware that it is important not to cause any disturbance to commercial services when evaluating real-world applications.
Prior to experiments with the Clarifai service, we received permission from Clarifai to use the public API with a certain budget for the research purpose.

\subsubsection{Attack Results}

As illustrated in Figure~\ref{fig:figure_7}, our method is suitable for evaluating real-world black-box applications' robustness.
We demonstrate that our method requires a feasible number of queries (about 20,000) with a specific source image.
We note that the targeted attack for this real-world application is challenging because this service contains more than 10,000 identities.
When running our attack method 5 times with random celebrity image pairs ($x_{src}$, $x_{trg}$), we have gotten the average value of SIM is 0.694 and the average value of LPIPS is 0.125.
As a result, the Clarifai celebrity recognition service shows better robustness than our trained facial identity recognition model.
\section{Discussion}

We have found a generative model can efficiently create various images that are classified as a specific target class if the distributions of both a generative model and a classification model are similar to each other.
We present a brute-forcing method for finding numerous target images.
For validating the brute-forcing method, we simply use a pre-trained StyleGAN model \cite{karras2019style}.
We have demonstrated that sampling random image $G(w)$ with a large number of queries can find various images that could be classified as a target class (Table~\ref{table_10}).
For example, the brute-forcing with 80,000 queries can find the majority classes of the facial identity recognition model including high confidence ($>90\%$) images.
We note that the instance generated by brute-forcing is not appropriate as an adversarial example itself because this instance may have similar semantic features to the target class images.
However, our result shows that if the GAN learns a similar distribution of the classification model and has enough capability for sampling various images, it is easy to find various images that are classified as a specific target class.
For example, we can use a found image by brute-forcing as an initial starting point in our attack procedure.
\section{Conclusion}

In this work, we present Latent-HSJA, a novel method for generating unrestricted adversarial examples in a black-box setting.
We demonstrate that our method can successfully attack state-of-the-art classification models, including a real-world application.
Our method can explore the latent space and generate various realistic adversarial examples in terms of that the latent space of GAN contains a large number of adversarial examples.
We especially utilize the valuable features of the StyleGAN2 architecture that have highly disentangled latent representations for a black-box attack.
The experimental results show that our attack method has the potential as a new adversarial attack method.
However, we have observed that the adversarial example is sometimes not feasible even though $MSE(w_{adv}, w_{src})$ is small enough. 
Therefore, future work may seek a better evaluation metric useful for our unrestricted adversarial attack.
In addition, for generating strong a adversarial example, an attacker can use a hybrid approach that combines our Latent-HSJA with other norm-based decision-based attacks in terms of that our attack method shows better results in an early stage of the attack.
Although we focus on the unrestricted attack on the face-related applications in this work, we believe our method is scalable to other classification tasks because GAN networks can be trained on various datasets.
We hope our work has demonstrated new possibilities for generating semantic adversarial examples in a real-world black-box scenario.

\clearpage
%
%
\bibliographystyle{splncs04}
\bibliography{egbib}
\end{document}